\documentclass[letterpaper, 10 pt, conference]{ieeeconf}  

\IEEEoverridecommandlockouts                              

\usepackage{color}
\usepackage{amsmath,amsfonts}
\usepackage{algorithm}
\usepackage{algorithmicx}
\usepackage{algpseudocode}

\usepackage{array}
\usepackage{booktabs}
\usepackage{textcomp}
\usepackage{stfloats}
\usepackage{url}
\usepackage{verbatim}
\usepackage{graphicx}
\usepackage{cite}
\usepackage{color, soul, framed}
\usepackage{makecell}
\usepackage{multirow}
\usepackage{wrapfig}
\usepackage{lipsum}
\usepackage{caption}
\usepackage{subcaption}
\usepackage{utfsym}
\usepackage{threeparttable}

\title{\LARGE \bf
Zero-Shot Object Goal Visual Navigation
}

\author{Qianfan Zhao$^{1}$, Lu Zhang$^{1}$, Bin He$^{3}$, Hong Qiao$^{1}$, and Zhiyong Liu$^{1,2}$
\thanks{}
\thanks{$^{1}$Qianfan Zhao, Lu Zhang, Hong Qiao, and Zhiyong Liu are with the State Key Laboratory of Management and Control for Complex Systems, Institute of Automation, Chinese Academy of Sciences, Beijing 100190, China, and also with the School of Artificial Intelligence, University of Chinese Academy of Sciences, Beijing 100190, China.
        {\tt\small zhaoqianfan2019@ia.ac.cn}, {\tt\small zhiyong.liu@ia.ac.cn}}%
\thanks{$^{2}$Zhiyong Liu is also with the Nanjing Artificial Intelligence Research of IA, Jiangning District, Nanjing, 211100, Jiangsu, China.
        }%
\thanks{$^{3}$Bin He is with the College of Electronic and Information Engineering, Tongji University, China.
        }%
}

\begin{document}

\maketitle
\thispagestyle{empty}
\pagestyle{empty}

\begin{abstract}

Object goal visual navigation is a challenging task that aims to guide a robot to find the target object based on its visual observation, and the target is limited to the classes pre-defined in the training stage. However, in real households, there may exist numerous target classes that the robot needs to deal with, and it is hard for all of these classes to be contained in the training stage. To address this challenge, we study the zero-shot object goal visual navigation task, which aims at guiding robots to find targets belonging to novel classes without any training samples. To this end, we also propose a novel zero-shot object navigation framework called semantic similarity network (SSNet). Our framework use the detection results and the cosine similarity between semantic word embeddings as input. Such type of input data has a weak correlation with classes and thus our framework has the ability to generalize the policy to novel classes. Extensive experiments on the AI2-THOR platform show that our model outperforms the baseline models in the zero-shot object navigation task, which proves the generalization ability of our model. Our code is available at: \url{https://github.com/pioneer-innovation/Zero-Shot-Object-Navigation}.

\end{abstract}

\section{INTRODUCTION}

Object goal visual navigation is an important skill for robots to perform real-world tasks, which seeks to guide the robot to reach the target based on its visual observations. Learning effective navigation policy is a complicated problem that involves many fields of robotics such as visual perception, scene understanding, and motion planning. While researchers have achieved promising object navigation results~\cite{article1,article2,article3,article4,article5}, these methods mainly focuses on the classes pre-defined in the training phase, which are called "seen classes". However, in real households, there may exist numerous object classes that cannot be fully contained in the training stage, which are called "unseen classes".

Therefore, inspired by the recent success of zero-shot learning in computer vision (e.g., image recognition~\cite{article11,article12}, object detection~\cite{article9,article10,article13}, and semantic segmentation~\cite{article14,article15}), we study the Zero-Shot Object Goal Visual Navigation (ZSON) task which intends to enable the robot to navigate to the target of unseen classes.

\begin{figure*}[thpb]
\centering
\includegraphics[width=5.8in]{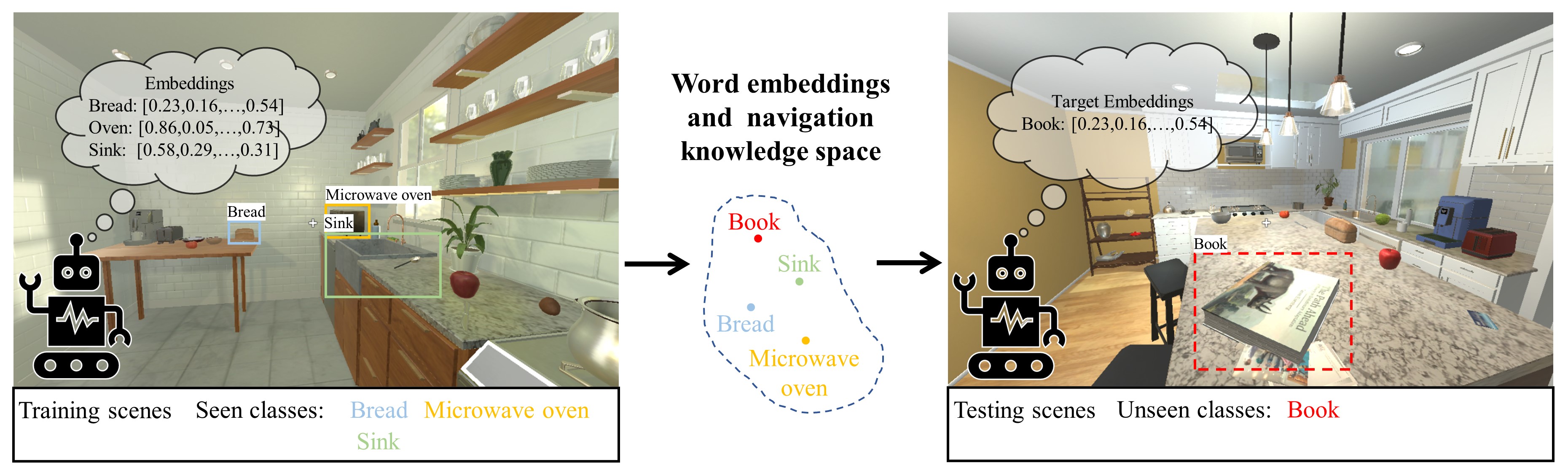}
\caption{The visualization of an example of the Zero-Shot Object Goal Visual Navigation task. The robot learns navigation policy in training scenes with observations and semantic embeddings of specified seen categories. The robots use the semantic embeddings of unseen categories to generalize the navigation policy, and successfully find the target object of the unseen category in testing scenes.}
\label{fig_1}
\end{figure*}

Zero-shot learning methods mainly uses images and semantic embeddings to align the visual features and semantic knowledge of seen classes into a common space in the training stage, and uses the semantic embeddings of unseen classes and the aligned common space to handle unseen classes in the testing stage~\cite{article11,article12,article9,article10,article13,article14,article15}. The ZSON task has a similar definition to the zero-shot learning. A concise example of ZSON task is illustrated in Fig. \ref{fig_1}. In the training stage, an agent is trained through visual features and semantic embeddings of seen classes in the training scenes. In the testing stage, the agent takes the observations in the testing scenes and the semantic embedding of the unseen class as input and output motion plans to find the unseen target. To successfully complete the task, the agent needs to get close enough to the target and adjust its observation angle until the target is visible.

Furthermore, we propose the semantic similarity network (SSNet) based on the deep reinforcement learning (DRL) algorithm and the self-attention mechanism. We believe that the overfitting of the model to the seen classes in the training stage is an important factor that prevents the model from being able to handle unseen classes. Therefore, we discard the widely used input data with strong correlation to classes, such as dense visual features and knowledge graphs~\cite{article1,article2,article3,article4,article5}, and utilize the detection results and the cosine similarity of semantic embeddings, which eliminates most of the information directly related to classes. With this approach, our model can avoid the overfitting on seen classes and generalize the learned policy to unseen classes. In addition, we also propose a new semantic reward function, which uses the cosine similarity of semantic embeddings to help the agent to learn navigation policy. This reward encourages the robot to look for objects semantically similar to the target during the navigation, and the target is more likely to appear around these similar objects.

We evaluate our method in the widely used platform AI2-THOR~\cite{article1} , which contains 120 simulated indoor scenes and 22 commonly used target classes~\cite{article3,article4,article5}. We re-split the 22 target classes into different numbers of seen classes and unseen classes. The experimental results confirm our view on the relationship between input data and generalization ability. The proposed model outperforms all baseline models on both seen and unseen classes. Furthermore, our model also shows considerable performance under the normal object navigation setting. 

In summary, we study the zero-shot object goal visual navigation task to solve the need for robots to navigate to unseen classes of targets in the open world. A novel zero-shot object navigation model called SSNet is proposed by us. This model is based on the DRL algorithm and use class-unrelated data as input, which can generalize the learned navigation policy to unseen classes. A novel semantic reward function is proposed by us. This reward function uses the cosine similarity of semantic embeddings between the target and other visible objects to guide the robot to learn the zero-shot object navigation policy.






\section{Related Work}
This study is related to visual navigation and zero-shot learning, which are briefly discussed as below.

\subsection{Visual Navigation}

As an important robotic task, visual navigation has attracted a lot of attention for a long time. After years of study, there is a number of works on visual navigation and we make a brief overview. Traditional navigation methods always rely on offline or online maps~\cite{article16,article17,article18,article19,article20,article21} made through Simultaneous Localization and Mapping (SLAM) techniques. They treat this task as an obstacle avoidance problem and focus on the path planning algorithm.

Recently, with the development of deep reinforcement learning, there are more advanced tasks are proposed in the field of visual navigation. According to different input and target types, recent visual navigation tasks can be divided into point goal visual navigation~\cite{article22,article23,article24}, object goal visual navigation~\cite{article1,article2,article3,article4,article5}, and vision-language navigation~\cite{article25,article26}. Our work is similar to object goal visual navigation, so we will only introduce the related work of object goal visual navigation. Object goal visual navigation refers to the task that the robot needs to learn a navigation policy to find a specified target instance and avoid obstacle collision. Scene-prior~\cite{article2} uses a knowledge graph to extract semantic priors and relationships of objects to navigate the agent. It uses a Graph Convolutional Network (GCN)~\cite{article28} to extract prior knowledge through the Visual Genome dataset~\cite{article27}. MJOLNIR~\cite{article4} uses a hierarchical object relationship reward, a context matrix, and a GCN to learn a navigation policy. VTNet~\cite{article5} uses a transformer to extract relationships among objects and set up strong connections with the navigation policy. The target-driven work~\cite{article1} uses an image of the target object as input to train navigation policy and build the widely used AI2-THOR framework which provides an environment with realistic 3D scenes and a physics engine. Although this line of work achieves promising results in visual navigation, they are limited in the classes specified in the training stage. Once they encounter targets that belong to novel classes in the testing stage, it is hard for them to complete their visual navigation tasks. Our work focuses on the zero-shot setting and achieves better results when dealing with unseen targets in test scenes.

\subsection{Zero-Shot Learning}

Zero-shot learning aims to use semantic embeddings (Word2vec~\cite{article30} or GloVe~\cite{article29}) to handle the unseen classes. In the early years, zero-shot learning research mainly focused on the classification problem~\cite{article11,article12}. With the emergence and development of other computer vision tasks, zero-shot learning has also been valued and applied by other field such as object detection~\cite{article9,article10,article13}, image segmentation~\cite{article14,article15}, etc. Zero-shot learning methods can be divided into two main categories: projection methods and generation methods. The projection methods project visual features and semantic embeddings of seen classes into a common space and align them by categories~\cite{article9,article12,article13,article14,article15}. When dealing with the unseen classes, their semantic embeddings can be used to infer their visual features in the common space. The generation methods~\cite{article10,article11} use the semantic embeddings and visual features of seen classes to train a generative model such as Generative Adversarial Networks (GAN)~\cite{article31} or Conditional Variational Autoencoders (CVAE)~\cite{article32}. The generative model can be used to generate samples of unseen classes through their semantic embeddings and the generated samples can be used to finetune the visual model. Our work is more similar to projection methods, but we do not use semantic embeddings as model input directly. Instead, we use the cosine similarity of the semantic embeddings between classes as the model input to generalize the navigation policy to unseen classes.




\section{Zero-Shot Object Navigation}

\subsection{Task Definition}

Consider a set of "seen" target classes denoted by $S = \{s_1,s_2,…,s_n\}$, which are available during the training process and $n$ stands for the total number of the seen classes. Consider another set of "unseen" target classes denoted by $U = \{u_1,u_2,…,u_m\}$, which are only available during the testing process and $m$ stands for the total number of unseen classes. Consider another set of "Irrelevant" classes denoted by $I = \{i_1,i_2,…,i_k\}$ and $k$ stands for the total number of Irrelevant classes, which exists in training scenes and testing scenes, but will not be selected as a target. In addition, we denote the set of "all" target classes by $C = S \cup U$. The set of all training scenes is denoted by $X_s$ and the set of test scenes is denoted by $X_c$. We also provide semantic embeddings (e.g., GloVe~\cite{article29} or Word2vec~\cite{article30}) $E_s$ for seen classes, $E_u$ for unseen classes, and $E_i$ for Irrelevant classes.

At the training phase, the agent is trained in training scenes $X_s$ with the seen target classes $S$. The agent is given visual observations $s$ of training scenes and semantic embeddings $E_s$ of the target to learn a navigation policy $\pi(s, E_s)$. During the testing phase, the agent is given visual observations of test scenes and semantic embeddings of the target belonging to all target classes $C$. When dealing with the target belonging to unseen classes $U$, the agent needs to generalize the learned navigation policy to the target using $E_u$, denoted as $\pi(s, E_u)$. Like most object navigation works~\cite{article1,article2,article3,article4,article5}, the zero-shot object navigation task is considered successful when the target object is visible in the current observation and within a threshold of distance (1.5m).

\subsection{Model Architecture}

\begin{figure*}[thp]
\centering
\includegraphics[width=5.7in]{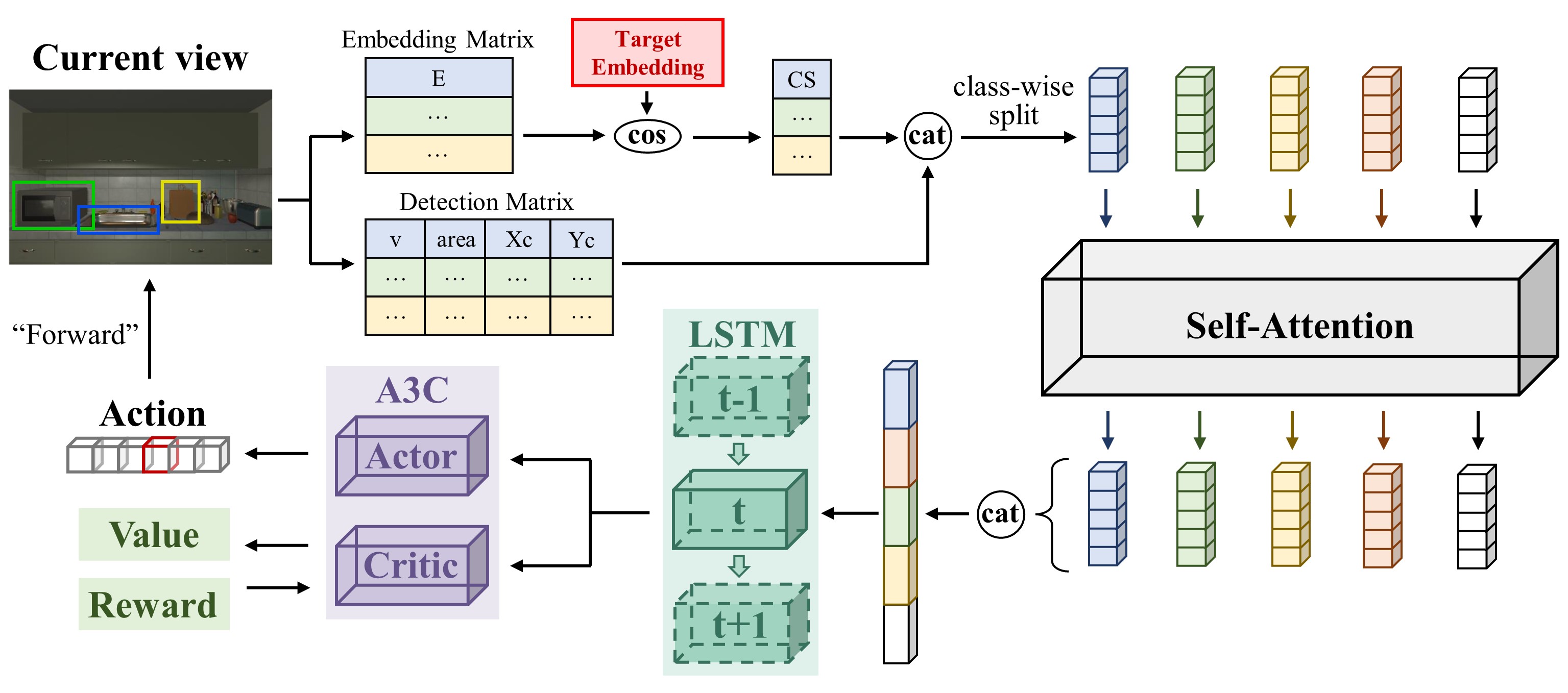}
\caption{The visualization of the proposed SSNet architecture. The input of the model is only the detection result and cosine similarity of semantic embeddings. The input is split along the category axis and processed by the self-attention module. The outputs of the self-attention module are concatenated into a 1-D vector. Then a LSTM module is used to extract and store the information of previous actions and an A3C module is used to generate actions.}
\label{fig_2}
\end{figure*}

The architecture of the proposed SSNet is shown in Fig. \ref{fig_2}. Different from other navigation works, we discard commonly used visual features or knowledge graphs~\cite{article1,article2,article4}, and only use a highly abstract detection results matrix and the cosine similarity of semantic embeddings as input. Inspired by~\cite{article4}, the detection matrix contains detection results of seen classes and irrelevant classes in the current observation (Because our work mainly focuses on the policy learning, we directly use the ground truth detection results as a perfect detector, which is the same as other papers~\cite{article3,article4,article33}). Each line of the detection matrix can be represented as $c_j = [v, x_c, y_c, area]$ of class $j$. The first element $v$ is of binary type. If the object of the class $j$ is visible in the current observation, the value of $v$ is one. If it cannot be observed, the value is zero. The second and third elements are the center coordinate of the detection bounding box of class $j$. The last element $area$ represents the image space of the bounding box. If no object of a class is visible, the variable values of the class in the detection matrix will be set to zero. In addition, there is an embedding matrix that represents the semantic embeddings of seen classes and irrelevant classes which are obtained from GloVe~\cite{article29}. the embedding matrix is used to calculate the cosine similarity (CS) with the semantic embedding of the target, which can be represented as below:

\begin{equation}
CosineSimilarity(g_j,g_t) = \frac{g_j \cdot g_t}{|g_j| * |g_t|}, \label{fm_1}
\end{equation}

where $g_j$ denotes the semantic embeddings of class $j$ and $g_t$ denotes the semantic embedding of the target. Finally, the CS will be concatenated with the detection matrix as input to the subsequent modules. It can be seen from the above description that our input rarely contains information directly related to the class, and most of the information is similarity value and detection bounding box parameters, which are class-unrelated.

After building the matrix, we introduce a self-attention module and use the concatenated matrix, which is split along the class axis, as the input. The self-attention module can adaptively learn the relationship between each class according to the different CS and detection results. The self-attention module uses the learned attention parameters to fuse and output features for each class. Then we concatenate the output features into a 1-D vector and use a Long Short Term Memory (LSTM) network~\cite{article34} to extract and store useful information from previous and current states. After the LSTM network, we adapt A3C algorithm~\cite{article35} to learn the visual navigation policy and output motion plans.

After sufficient training, our model can learn the relationship between different classes to facilitate visual navigation policy. In the testing phase, the testing scenes contain seen and unseen classes that can be selected as the target. When dealing with unseen targets, our model only needs to take the semantic embedding of the target to calculate its cosine similarity with the embedding matrix, and then it can generalize the learned navigation policy to unseen classes without any other measures.

\subsection{Learning Set-Up}

After introducing the model architecture, we will describe the reinforcement learning settings: action space, observations, and reward.

\textbf{Action space :}
We use the same discrete action space as in other papers~\cite{article1,article2,article4,article5} when simulating on a virtual platform (AI2-THOR). The discrete action space $A = \{MoveAhead,\ RotateLeft,\ RotateRight,\ LookUp,\\ LookDown,\ Done\}$. The $MoveAhead$ action will move the robot forward 0.25 meters. The $RotateLeft$ and $RotateRight$ action will rotate the robot 45 degrees. The $LookUp$ or $LookDown$ action will tilt the camera up or down by 30 degrees. The $Done$ action represents that the robot believes it has found the target and the episode will end.

\textbf{Observations :}
The robot in the AI2-THOR framework will take RGB images as observations.

\textbf{Reward :}
We propose a novel semantic reward to utilize the cosine similarity of semantic embeddings to learn visual navigation policy. The reward value is obtained by calculating the cosine similarity of the semantic embeddings between the visible objects and the target. If there are multiple objects that can be observed at the current observation, we choose the max cosine similarity value as the reward. In this way, it can encourage the agent to find the objects which are semantically similar to the target. Since the entire model is trained end-to-end, the reward will propagate back to the self-attention module, and guide it to correctly learn the attention parameters between the objects. In addition, only when the current similarity value is greater than the last similarity value will it be used as a reward value. This encourages the robot to find objects with higher semantic similarity until the target is located. Finally, if there are no objects visible in the current observation, the robot will get a reward value of -0.01 as a penalty to reduce the trajectory length. The above reward calculation process can be summarized by Algorithm~\ref{algorithm1}.

\begin{algorithm}[th]
  \caption{Semantic Reward}
  \label{algorithm1}
  \begin{algorithmic}[1]
   	\Require $s$ state, $a$ action, $t$ target, $obj$ objects, $CS_{max}=0$, $SC$ SeenClasses, $IC$ IrrelevantClasses,
	\Ensure $r$ reward
	\Function {S-Reward}{$s$,$a$,$t$,$obj$,$CS_{max}$,$SC$,$IC$} 
	\If {$a = Done$}
		\If {$obj_i$ is visible}
			\State $r = 5$	
		\Else
			\State $r = 0$	
		\EndIf
		\State \Return $r$
	\Else
		\State $r = - 0.01$	
		\For {$obj_i\in$\{$SC$\}$\cup$\{$IC$\}}	
			\If {$obj_i$ is visible}
				\State $CS = CosineSimilarity(obj_i,t)$	
				\If {$CS \textgreater CS_{max}$}  
					\State $r = CS$		
					\State $CS_{max} = CS$
				\EndIf
			\EndIf
		\EndFor
		\State \Return $r$
	\EndIf
	\EndFunction
  \end{algorithmic}
\end{algorithm}


\section{Experiment and Result}

\subsection{Experiment Setting}

We use the AI2-THOR embodied AI environment as the platform for the zero-shot object navigation task. This environment contains 120 photo-realistic floorplans including 4 different room layouts: Kitchen, Livingroom, Bedroom, and Bathroom. Each room contains a number of objects that the agent can observe and interact with. Similar to other papers~\cite{article3,article4,article5}, we also use 80 rooms as the training scenes and use 40 rooms as the testing scenes. We re-split the widely used 22 target classes~\cite{article3,article4,article5} into 18/4 seen/unseen and 14/8 seen/unseen classes, and evaluate models on both class splits.

\subsection{Baseline Models}

\textbf{Random :} the robot randomly selects actions from its action space.

\textbf{MJOLNIR :} we train this off-shelf object navigation model~\cite{article4} under the zero-shot object navigation setting. This model uses a GCN stream and an observation stream to extract context information to learn navigation policy.

\begin{figure}[thpb]
\centering
\includegraphics[width=2.8in]{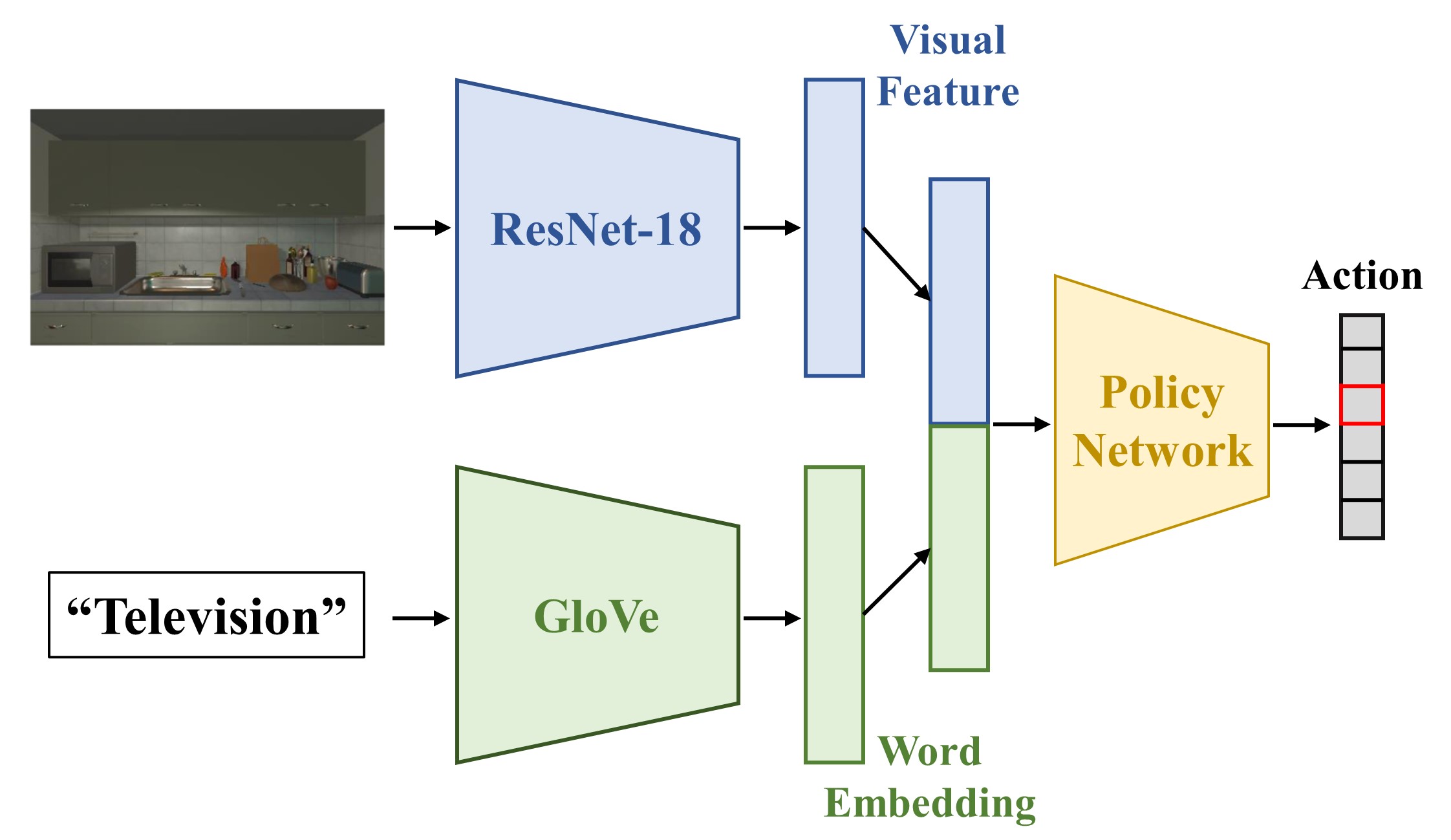}
\caption{The visualization of the ZS-Baseline model architecture.}
\label{fig_3}
\end{figure}

\textbf{Zero-Shot Baseline (ZS-Baseline) :} we build another zero-shot navigation baseline model. Its architecture is shown in Fig. \ref{fig_3}. We directly use a pre-trained ResNet to extract a 1-D visual feature from the current observation and use the semantic embedding of the target class to concatenate with the visual feature as the input of the policy network. The policy network is composed of LSTM and A3C, which is similar to the proposed model in section 3.2.

\subsection{Metrics}
In order to facilitate and fair comparison with baseline models, we use the evaluation metrics proposed by~\cite{article36} which are also widely used in other object navigation algorithms~\cite{article3,article4,article5}. These metrics contain Success Rate (SR) and Success weighted by Path Length (SPL). The SR is defined as $\frac{1}{N}\sum_{i=1}^n S_i$ , and the SPL is defined as $\frac{1}{N}\sum_{i=1}^n S_i \frac{l_i}{max(l_i,e_i)}$. $N$ is the number of episodes. $S_i$ is a binary vector indicating the success of the $i$-th episode. $e_i$ is the path length of the agent in an episode. $l_i$ is the length of the optimal path to the target. We divide the paths planned by the model into two categories: one is that the path length is greater than 1 (L\textgreater{}=1), and the other is that the path length is greater than 5 (L\textgreater{}=5). Consistent with most algorithms~\cite{article3,article4,article5}, we evaluate two path paths of different lengths separately.

\subsection{Implementation Details}
We build our models based on the open-source code of~\cite{article4,article3}. All models were trained for 900,000 episodes on the offline data from AI2-THOR~\cite{article1} with a 0.0001 learning rate. During the evaluation, we evaluate 250 episodes for each room type. In each evaluation episode, the floorplan, initial position, and target are randomly chosen.

\subsection{Experimental Results}

\renewcommand{\arraystretch}{1.7}
\begin{table*}[thpb]
\centering
\caption{Performances of the baseline models and ours in testing scenes.}
\label{tab:table1}
\begin{tabular}{cccclcclcclcc}
\bottomrule[1.2pt]
\specialrule{0em}{0.5pt}{0.5pt}
\bottomrule[0.5pt]
\multirow{3}{*}{Model} & \multirow{3}{*}{\begin{tabular}[c]{@{}c@{}}Seen/Unseen\\ split\end{tabular}} & \multicolumn{5}{c}{Unseen classes}                                               &  & \multicolumn{5}{c}{Seen classes}                                                 \\ \cline{3-7} \cline{9-13} 
                       &                                                                              & \multicolumn{2}{c}{L\textgreater{}=1} &  & \multicolumn{2}{c}{L\textgreater{}=5} &  & \multicolumn{2}{c}{L\textgreater{}=1} &  & \multicolumn{2}{c}{L\textgreater{}=5} \\ \cline{3-4} \cline{6-7} \cline{9-10} \cline{12-13} 
                       &                                                                              & SR(\%)             & SPL(\%)          &  & SR(\%)            & SPL(\%)           &  & SR(\%)            & SPL(\%)           &  & SR(\%)            & SPL(\%)           \\ \hline
Random                 & 18/4                                                                         & 10.8               & 2.1              &  & 0.9               & 0.3               &  & 9.5              & 3.3               &  & 1.0               & 0.4               \\
ZS-Baseline            & 18/4                                                                         & 16.9               & 8.7              &  & 5.3               & 3.1               &  & 17.7              & 8.3               &  & 5.3               & 2.6               \\
MJOLNIR~\cite{article4}                & 18/4                                                                         & 20.7               & 7.1              &  & 10.6               & 4.5               &  & 51.9              & 16.5              &  & 33.0              & 14.2              \\ \hline
Ours                   & 18/4                                                                         & \textbf{28.6}      & \textbf{9.0}     &  & \textbf{12.5}      & \textbf{5.6}      &  & \textbf{59.0}     & \textbf{19.7}     &  & \textbf{38.6}     & \textbf{18.3}     \\ \bottomrule[1pt]
Random                 & 14/8                                                                         & 8.2               & 3.5              &  & 0.5               & 0.1               &  & 8.9              & 3.0             &  & 0.5               & 0.3              \\
ZS-Baseline            & 14/8                                                                         & 14.6               & 4.9              &  & 4.9               & 2.8               &  & 30.4              & 9.7               &  & 11.5               & 5.2               \\
MJOLNIR~\cite{article4}                & 14/8                                                                         & 12.3               & 5.1              &  & 6.0               & 3.6               &  & 52.7              & 22.3              &  & 26.8              & 14.9              \\ \hline
Ours                   & 14/8                                                                         & \textbf{21.5}      & \textbf{7.0}     &  & \textbf{13.0}      & \textbf{6.7}      &  & \textbf{59.3}     & \textbf{24.5}     &  & \textbf{35.2}     & \textbf{19.3}     \\
\bottomrule[0.5pt]
\specialrule{0em}{0.5pt}{0.5pt}
\bottomrule[1.2pt]
\end{tabular}
\end{table*}

Table~\ref{tab:table1} shows the performance of baseline models and ours in testing scenes under different unseen class splits. The testing scenes contain the rooms which did not appear during the training phase, and thus the locations of the objects in the testing scenes are completely unknown. It can be seen that our model outperforms all the baseline models in unseen classes. These results prove that using the detection results and the cosine similarity of semantic embeddings allows the model to better generalize the navigation policy to unseen classes, and the self-attention module can better extract useful information for navigation policy. Under different split settings, our model can still be effective, but the success rate decreases when the number of unseen classes increases. We note that as the number of seen classes decreases, it will be difficult for the model to learn the relationship between semantic knowledge and the navigation policy, resulting in the reduction of generalization ability.


\begin{figure*}[thpb]
\centering
\includegraphics[width=5.5in]{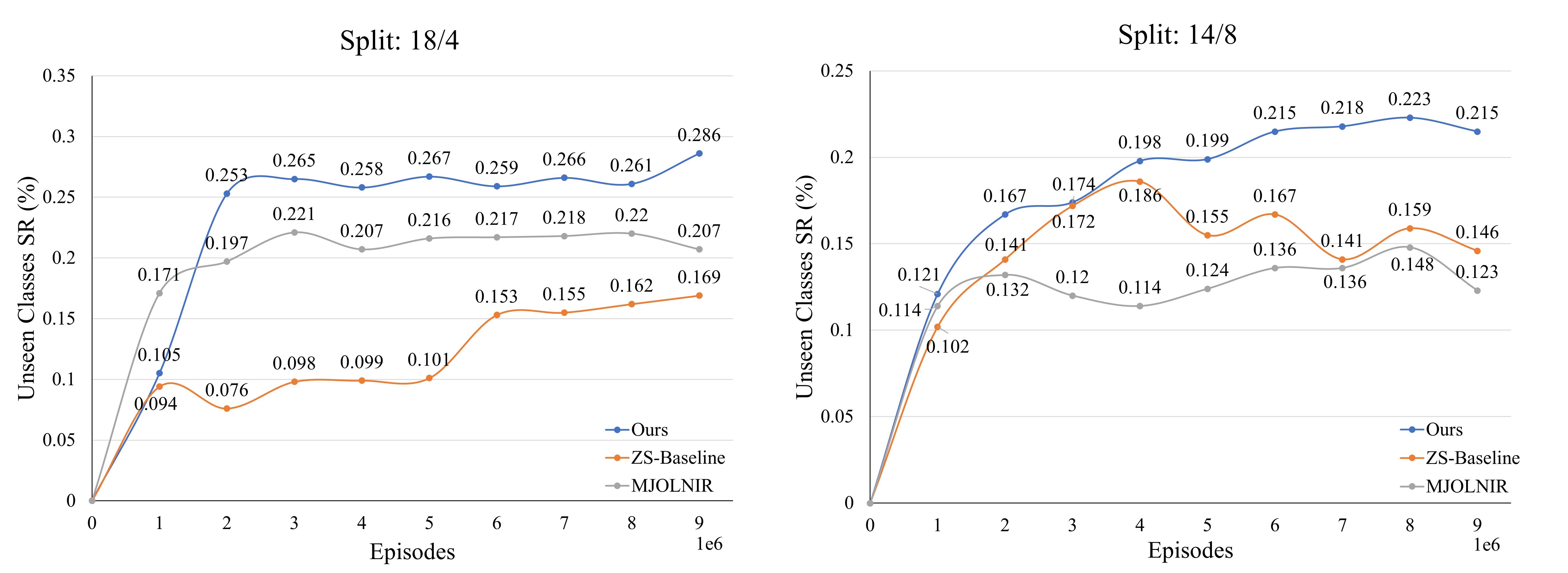}
\caption{The visualization of success rate (L\textgreater{}=1) of different models under different class splits.}
\label{fig_4}
\end{figure*}

Fig. \ref{fig_4} shows the success rate for some baseline models and ours. Our model achieves a better success rate as the episode grows. In addition, our model also learns faster than other models, which can be proved by the faster growth of the success rate. It is worth noting that the baseline models achieve a better performance at the beginning of the training stage and gradually degrade in subsequent training processes, while the performance of our model grows steadily. We believe that this phenomenon is caused by the overfitting of end-to-end learning and redundant inputs. This phenomenon also proves that using abstract input can reduce overfitting and enhance generalization ability.

\begin{table}
\centering
\caption{Performances of the models under the normal setting.}
\label{tab:table2}
\begin{tabular}{cccccc}
\bottomrule[1pt]
\multirow{2}{*}{Model} & \multicolumn{2}{c}{L\textgreater{}=1} &  & \multicolumn{2}{c}{L\textgreater{}=5} \\ \cline{2-3} \cline{5-6} 
                        & SR         & SPL       &  & SR        & SPL               \\ \hline
ZS-Baseline             & 18.7       & 7.9       &  & 4.3       & 2.3             \\
MJOLNIR~\cite{article4} & \textbf{65.3}       & 21.1      &  & \textbf{50.0}      & 20.9              \\ \hline
Ours                    & 63.7       & \textbf{22.8}      &  & 42.9      & \textbf{21.3}    \\ \bottomrule[1pt]
\end{tabular}
\end{table}

Table~\ref{tab:table2} shows the performance of models under the normal object navigation setting. In the case of better ZSON performance of our model,  compared with other baseline models, our method is still effective and achieves better SPL.

Table~\ref{tab:table3} shows the ablation study of our model. In this study, we mainly focus on two major modules: self-attention (SA) and partial reward (PR). We first remove the both modules and the performance drops by about 30$\%$. Then we only remove the partial reward, the performance increases by about 35$\%$, which shows that the self-attention modules can indeed help in processing detection matrix and cosine similarity of semantic embeddings. Finally, we add the partial reward function and the performance increases by about 5$\%$. Our partial reward improve the performance a little by providing denser rewards.

\begin{table}
\centering
\caption{Ablation study of our model under the 18/4 zero-shot setting.}
\label{tab:table3}
\begin{threeparttable}
\begin{tabular}{cclccccc}
\bottomrule[1pt]
\multicolumn{2}{c}{Module} &  & \multicolumn{2}{c}{L\textgreater{}=1} &  & \multicolumn{2}{c}{L\textgreater{}=5} \\ \cline{1-2} \cline{4-5} \cline{7-8} 
SA\tnote{1}  & PR\tnote{2}   &  & SR                & SPL               &  & SR                & SPL               \\ \hline
$\usym{2613}$& $\usym{2613}$ &  & 20.3              & 4.5               &  & 10.7              & 4.3              \\
$\checkmark$ & $\usym{2613}$ &  & 27.9              & 8.2               &  & 12.1              & 4.8             \\
$\checkmark$ & $\checkmark$  &  & 28.6              & 9.0               &  & 12.5              & 5.6              \\ \bottomrule[1pt]
\end{tabular}
\begin{tablenotes}
        \footnotesize
        \item[1] SA: self attention module.
	   \item[2] PR: partial reward function.
      \end{tablenotes}
\end{threeparttable}
\end{table}

\subsection{Limitation}
In this paper, we only conduct experiments on a embodied AI platform and do not verify on a physical robot. The room layout in AI-THOR is relatively simple, and there are no complex suite layouts. We use the ground truth instead of a object detector because we mainly focus on the navigation policy. Obviously, real houses have very complicated room layouts and require a more robust policy to guide robots. Future work will explore the zero-shot object navigation task in more complex room layouts and evaluate our policy on a physical robot.

\section{Conclusion and Future Work}

In this paper, we study the Zero-Shot Object Goal Visual Navigation task that aims to navigate the robot to a novel target in testing scenes and the navigation model could not learn semantic knowledge of unseen classes through image and semantic embedding samples in the training stage. We also proposed the semantic similarity network (SSNet) based on the DRL algorithm and the self-attention module. Our model utilizes class-unrelated data as input to alleviate the overfitting of seen classes. Extensive experiments show that our model can successfully generalize the learned navigation policy to unseen classes and significantly outperforms other baseline models in both seen and unseen classes.

Since the input of our model is highly abstract information, we expect that our model may have good sim-to-real transfer ability. In future work, we will apply our model to the physical robot platform and build a digital twin environment for sim-to-real transfer experiments. Furthermore, we will combine our model with the life-long learning method to explore the life-long learning capabilities of robots.

\section*{ACKNOWLEDGMENT}
This work is supported by National Key Research and Development Plan of China grant 2020AAA0108902, the Strategic Priority Research Program of Chinese Academy of Science under Grant XDB32050100, NSFC, China grants 61627808, FuJian Science and Technology Plan under Grant No. 2021T3003, and Dongguan core technology research frontier project (2019622101001).

\bibliographystyle{IEEEtran}
\bibliography{IEEEabrv,IEEEexample}

\end{document}